\documentclass[10pt,journal,compsoc]{IEEEtran}
\usepackage{amsmath,amsfonts}
\usepackage{algorithmic}
\usepackage{algorithm}
\usepackage{array}
\usepackage[caption=false,font=normalsize,labelfont=sf,textfont=sf]{subfig}
\usepackage{textcomp}
\usepackage{stfloats}
\usepackage{url}
\usepackage{verbatim}
\usepackage{graphicx}
\usepackage{cite}
\usepackage{hyperref}
\usepackage{url}
\usepackage{graphicx}
\usepackage{xcolor}
\usepackage{booktabs}
\usepackage{wrapfig}
\usepackage{amsfonts}       
\usepackage{nicefrac}       
\usepackage{microtype}      
\usepackage{xcolor}         
\usepackage{amsmath}
\usepackage{tabularx}

\begin{document}

\title{MMTryon: Multi-Modal Multi-Reference Control for High-Quality Fashion Generation}

\author{Xujie~Zhang$\ast$, 
        Ente~Lin$\ast$, 
        Michael~Kampffmeyer,
        Zhenyu~Xie,
        Jiang~Li,\\
        Ting~Liu,
        Xiaochao~Qu,
        Luoqi~Liu,
        Xiaodan~Liang$\dagger$,~\IEEEmembership{Senior Member,~IEEE}
        
\IEEEcompsocitemizethanks{\IEEEcompsocthanksitem X. Liang is the corresponding author. E-mail: xdliang328@gmail.com.\protect\\

\IEEEcompsocthanksitem Xujie Zhang and Ente Lin contributed equally to this work.
\IEEEcompsocthanksitem Xujie Zhang, Zhenyu~Xie, and Xiaodan Liang are with Shenzhen Campus of Sun Yat-sen University, China. Michael~Kampffmeyer is with UiT The Arctic University of Norway and the Norwegian Computing Center. Ente Lin is with Tsinghua University. Jiang Li, Ting Liu, Xiaochao Qu, and Luoqi Liu are with MT Lab.
}
}
\markboth{IEEE TRANSACTIONS ON PATTERN ANALYSIS AND MACHINE INTELLIGENCE, VOL.,No. XX, XX 2025}
{Shell \MakeLowercase{\textit{et al.}}: Bare Demo of IEEEtran.cls for Computer Society Journals}


\maketitle
\begin{abstract}
This paper introduces MMTryon, a multi-modal multi-reference VIrtual Try-ON (VITON) framework, which can generate high-quality compositional try-on results by taking a text instruction and multiple garment images as inputs. Our MMTryon addresses three problems overlooked in prior literature: 1) \textbf{Support of multiple try-on items.} Existing methods are commonly designed for single-item try-on tasks (e.g., upper/lower garments, dresses).
2) \textbf{Specification of dressing style}. Existing methods are unable to customize dressing styles based on instructions (e.g., zipped/unzipped, tuck-in/tuck-out, etc.)
3) \textbf{Segmentation Dependency}. They further heavily rely on category-specific segmentation models to identify the replacement regions, with segmentation errors directly leading to significant artifacts in the try-on results. To address the first two issues, our MMTryon introduces a novel multi-modality and multi-reference attention mechanism to combine the garment information from reference images and dressing-style information from text instructions. Besides, to remove the segmentation dependency, MMTryon uses a parsing-free garment encoder and leverages a novel scalable data generation pipeline to convert existing VITON datasets to a form that allows MMTryon to be trained without requiring any explicit segmentation. 
Extensive experiments on high-resolution benchmarks and in-the-wild test sets demonstrate MMTryon's superiority over existing SOTA methods both qualitatively and quantitatively. 
MMTryon's impressive performance on multi-item and style-controllable virtual try-on scenarios and its ability to try on any outfit in a large variety of scenarios from any source image, opens up a new avenue for future investigation in the fashion community. The project page is available at  \href{https://zhangxujie-gn.github.io/MMTryon.github.io/}{this link}.
\end{abstract}


\section{Introduction}
VIrtual Try-ON (VITON) technology aims to generate photo-realistic images featuring individuals adorned in specific garments according to input images of both the garments and the person. Due to its promising potential to revolutionize the online shopping experience, VITON has garnered considerable attention both in the academic and industry communities.
Despite the substantial progress achieved in academic benchmarks~\cite{choi2021viton,davide2022dress}, previous methods~\cite{xie2021vton,ge2021parser,he2022style,bai2022single,xie2023gp,morelli2023ladi,gou2023taming} still fall short of meeting the requirements of real-world applications, in particular for: 1) \textbf{Multiple-garment Compositional Try-On}: Users should be able to freely choose and combine a diverse range of tops, bottoms, and accessories from in-the-wild images, rather than being confined to the single-garment try-on scenario. 2) \textbf{Flexible Try-On Styles}: Users should have the flexibility to select their preferred try-on style, allowing them to specify how a garment is worn, such as deciding whether a top should be tucked into the trousers or draped over them.

Due to their focus on single garments, these methods encounter challenges when multiple garments and accessories need to be changed simultaneously. This limitation arises because collecting training data for single-garment try-on is relatively easy, whereas gathering data for multi-garment try-on is significantly more difficult. Additionally, these methods struggle to model the subtle differences in how each piece of clothing is styled during a compositional try-on, such as the overlapping relationship between tops and bottoms. Furthermore, given a flat in-store image of a coat with a zipper, existing methods are limited to replicating the garment in its original state onto the target, without the ability to adjust for open or closed styles. 

Additionally, existing methods lack the capability to precisely identify try-on areas, often resorting to trained segmentation models~\cite{Gong2019Graphonomy,li2020self} for explicit try-on area localization. This reliance on segmentation models makes these approaches highly dependent on the segmentation model's precision, which can be unsatisfactory in complex scenarios, introducing additional noise during training.
Additionally, conditioning the generation on the segmented garment results in a loss of information of how the garment is worn, leading to highly unrealistic try-on outcomes, particularly for uniquely designed garments. Moreover, the dependence on segmentation models severely restricts the flexibility of the try-on process, as these models are typically confined to a predefined set of garment classes.

\begin{figure*}[t]
  \centering
  \includegraphics[width=1.0\hsize]{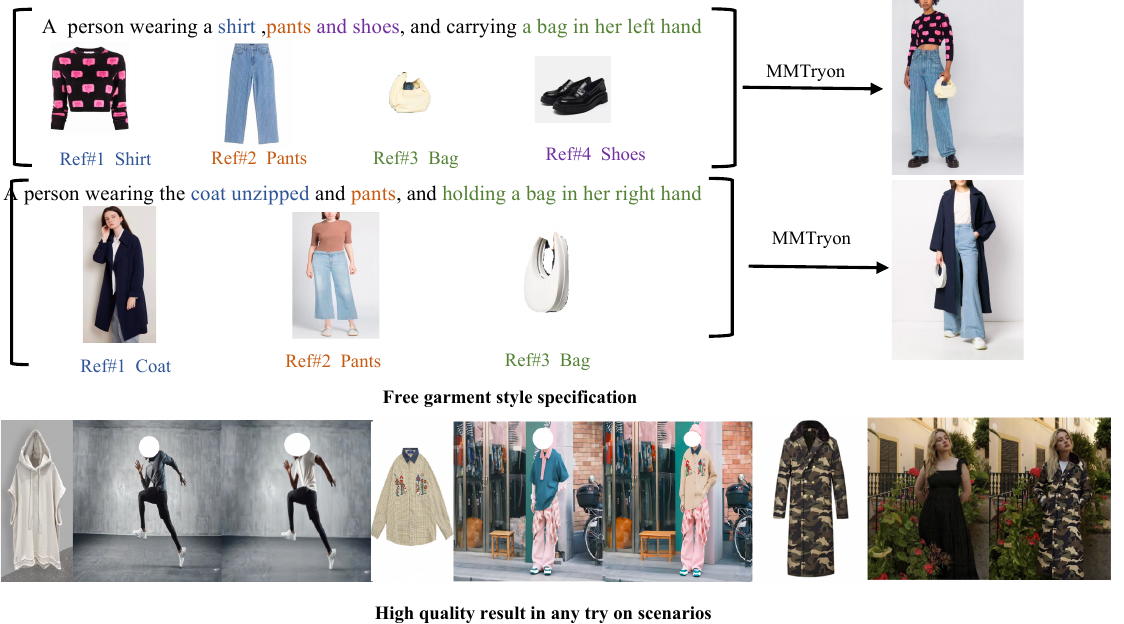}
  \caption{
   MMTryon can follow complex instructions to produce high-quality results in multi-garment/accessories compositional try-on scenarios. Besides, MMTryon can also generate high-fidelity results for any in-the-wild scenario.
   }
  \label{fig:teaser}
\end{figure*}
To address these challenges, we propose MMTryon, a more flexible try-on model capable of achieving multi-modal compositional dressing without the need for pre-segmenting the try-on areas. In our approach, a general try-on scenario is decoupled into text and image information, where the text information encompasses high-level garment descriptions and try-on style instructions, while the image information contains fine-grained texture details.
MMTryon is based on a conditional diffusion model that, to ensure efficient incorporation of both the image and text information in challenging scenarios, leverages our novel multi-reference image attention and multi-modal instruction attention modules. Specifically, to exploit the image information, we pre-trained a robust garment encoder on a large dataset to project image features into the multi-reference image attention module. This module then can utilize text as queries to extract the necessary features with the aim of eliminating our dependency on segmentation. For the text information, we develop a sophisticated multi-modal prompt and employ a pre-trained text encoder to project the text information into the specifically designed Multi-Modal Instruction Attention module. Finally, we train our model on a synthetically enhanced dataset, facilitated by our proposed data pipeline, using the pre-trained encoders and Stable Diffusion-initialized weights. Despite the difficulty of obtaining ideal multi-garment-to-single-model pairs for compositional try-on, our pre-trained garment encoder ensures excellent results even with limited datasets. Additionally, by overcoming the segmentation accuracy limitations in existing VITON models, our model can more effectively handle the nuances of different clothing items and styles (see Fig.~\ref{fig:teaser}).

Overall, our contributions can be summarized as follows: First, we propose MMTryon which, to our knowledge, is the first model to support the multi-modal compositional try-on task in a parsing-free manner. MMTryon enables  the combination of one or multiple garments for try-on, while also allowing the manipulation of try-on effects through textual commands. Notably, through our specially designed garment encoder and scalable data generation pipeline, MMTryon eliminates the need for prior segmentation networks during both training and inference, relying solely on text descriptions, individual garment images, or model images to identify areas of interest. Furthermore, to facilitate further research into this direction, we will publicly release a compositional try-on dataset with detailed annotations.

Finally, we demonstrate through extensive experiments on public try-on benchmarks and in-the-wild test sets that our approach outperforms existing state-of-the-art methods in both qualitative and quantitative evaluations.

\begin{figure*}
  \centering
  \includegraphics[width=1.0\textwidth]{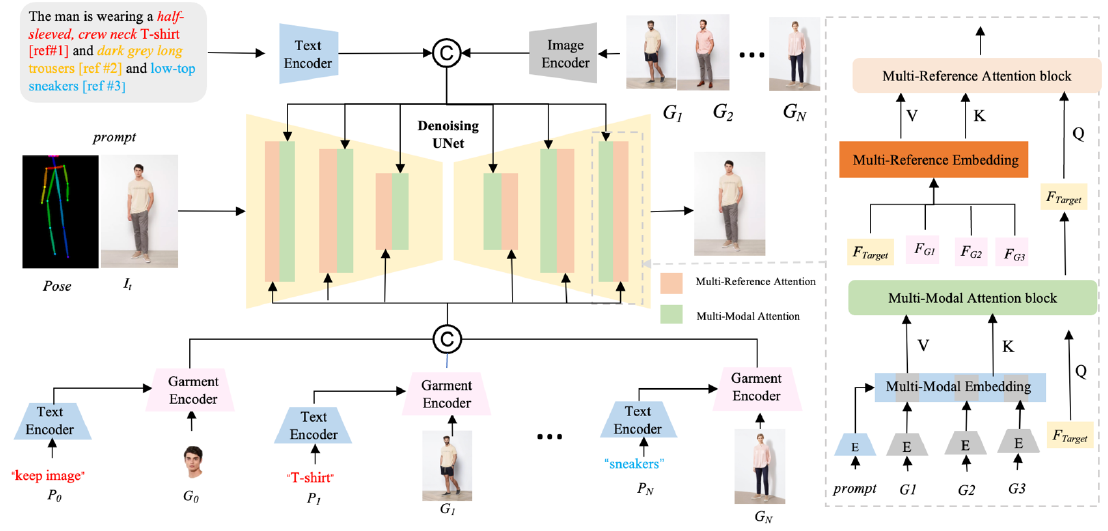}

  \caption{Overview of the proposed MMTryon framework. The instruction prompt and garment images are combined to obtain a multi-modal instruct embedding, replacing the original textual condition. Each garment image together with the corresponding text span are further processed by the garment encoder to obtain reference features, which, along with target features, undergo multi-reference attention to ensure detailed texture transfer. }
  \label{fig:framework}

\end{figure*}

\section{Related work}
\textbf{GAN-Based Virtual Try-on.}
Traditional virtual try-on methods~\cite{wang2018toward,dong2019towards,yang2020towards,ge2021parser,he2022style,choi2021viton,xie2021vton,zhao2021m3d,xie2021towards,lee2022high,bai2022single,davide2022dress,xie2022pasta,dong2022dressing,huang2022towards,xie2023gp} usually adopt a two-stage pipeline with Generative Adversarial Networks (GANs). The first stage employs an explicit warping module to deform in-shop clothing to the target shape, while the second stage uses a GAN-based generator to fuse the deformed clothing onto the target person. For these methods, the synthetic quality largely depends on the quality of the deformation in the first stage, prompting current methods to focus on enhancing the non-rigid deformation capabilities of the warping module. Meanwhile, some GAN-based approaches~\cite{cui2021dressing,xie2022pasta}  have explored the compositional try-on task, utilizing a cyclical generation mode to achieve sequential try-on. However, when the outcome of compositional dressing is entirely determined by the try-on order, the lack of understanding of the semantic information of the garments often leads to unrealistic results. 
Furthermore, most GAN-based solutions mentioned above directly utilize UNet-based generators for try-on synthesis, with little exploration into how to enhance the generator's capabilities. This results in poor resolution and visual quality of the try-on outcomes, making it difficult to integrate them with diverse linguistic instructions.

\noindent\textbf{Diffusion-Based Virtual Try-on.}
Compared to GAN-based models, diffusion models have made significant strides in high-fidelity conditional image generation~\cite{rombach2022high,saharia2022photorealistic,ramesh2022hierarchical}. Image-based virtual try-on is essentially a specialized form of the image editing/repair task conditioned on a given garment image. Therefore, a straightforward adaptation is to extend text-to-image diffusion models to accept images as conditions. While methods such as ~\cite{yang2023paint,huang2023composer,chen2023anydoor} have demonstrated their capabilities in virtual try-on, they fall short of preserving the texture details of the try-on results due to inadequate extraction of fine-grained features by the image encoder.
Recent methods~\cite{morelli2023ladi,gou2023taming} have further aimed to integrate traditional GAN-based approaches with diffusion models. They employ explicit warping modules to create deformed garments and use diffusion models to blend these with reference person images. Beyond simply incorporating explicit warping modules, the newly introduced TryonDiffusion~\cite{zhu2023tryondiffusion}, conversely, implements an implicit warping mechanism for clothing deformation. This addresses the issues of texture misalignment mentioned earlier and achieves promising try-on synthesis. However, TryonDiffusion relies on precise parsing to achieve various conditions, making it unable to handle more complex try-on scenarios. 
Recently, CatVTON~\cite{chong2025catvton} has further streamlined this process by eliminating heavy warping modules, achieving robust performance through efficient feature concatenation. Meanwhile, MV-VTON~\cite{wang2024mv} extends the scope of diffusion-based try-on to multi-view scenarios, ensuring consistency across different viewing angles. Despite these advancements, none of the existing approaches have effectively leveraged the capabilities of current large multi-modal models and have discarded text as an input, lacking the ability of free-form instruction following. The try-on effect in these approaches is solely inferred from the garment image.

Beyond single-garment transfer, recent research has focused on compositional and style-controllable tasks. MMVTON~\cite{mmvton} utilizes multiple garments as input but is limited to label-level text and requires explicit segmentation during inference. AnyFit~\cite{li2024anyfit} significantly scales up compositional try-on to high-resolution benchmarks, achieving robust fitting for diverse attire combinations. However, AnyFit treats garments primarily as static textures to be fitted and lacks the capability to customize dressing styles (e.g., ``zipped'' vs. ``unzipped'') via free-form instructions.
To enable fine-grained style control, Wear-Any-Way~\cite{chen2024wear} introduces a manipulable pipeline using sparse point-based interactions to guide the generation (e.g., dragging points to roll up sleeves). While effective, point-based control requires manual user intervention and is less intuitive/scalable than natural language commands. ImageDressing~\cite{shen2024imagdressing} employs lightweight adapter training for flexible generation but does not explicitly tackle the multi-reference consistency problem under complex instructions. 
In contrast to these approaches, our MMTryon effectively leverages the capabilities of large multi-modal models to support multi-modal compositional control. Unlike Wear-Any-Way's point interactions or AnyFit's fixed styles, MMTryon allows users to control both the garment combination and the dressing style via natural language. Crucially, MMTryon operates in a parsing-free manner during inference, eliminating the dependency on error-prone segmentation models. This not only addresses the limitations seen in both GAN and diffusion model-based methods but also utilizes the strengths of large multi-modal models, allowing for more detailed and flexible adaptation to various clothing styles and try-on scenarios.

\begin{figure}
  \centering
  \includegraphics[width=0.51\textwidth]{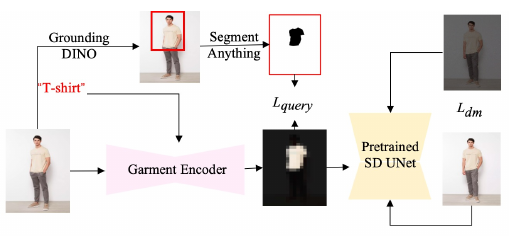}
\caption{Overview of the proposed pretrained garment encoder. Our garment encoder utilizes a prior mask derived from Grounding DINO and SAM to improve text query accuracy through cross-attention between the target text and the input features. The garment encoder is supervised by the diffusion reconstruction loss $L_{dm}$ and our text query loss $L_{query}$.}
  \label{fig:pretrain}

\end{figure}

\section{Method}\label{sec:method}

As shown in Fig.~\ref{fig:framework}, MMTryon has a minimal input setup during inference, only requiring a source person image, multiple reference images, and a text instruction describing how the items from the references are fitted onto the desired person. To achieve multi-modal multi-reference controllable try-on, two key elements are central: the training data and the model design. 
We start by presenting the design of our model to support this inference setup before describing how the data is gathered to support model training.
\vspace{-2mm}

\subsection{Overview}
Given a target person image $I$, multiple reference images containing try-on items $\{G_i\}$, and a text instruction $T$ describing how the items from those reference images are dressed, MMTryon aims to seamlessly follow $T$ to synthesize photo-realistic try-on results $I'$. We model this task as a conditional diffusion model, which takes $I$, $\{G_i\}$, and $T$ as conditions and gradually estimates a denoised version of $I'$ from pure noise. Our diffusion model is built upon the publicly available Stable Diffusion text-to-image model~\cite{rombach2022high}.


Stable Diffusion is a  text conditioned latent diffusion model~\cite{rombach2022high}. 
For an VAE~\cite{kingma2013auto} encoded image latent feature $z_0$, the forward diffusion process is performed by adding noise according to a predefined noise scheduler $\alpha_t$\cite{ho2020denoising}:
\begin{equation}
q(z_t|z_0) = \mathcal{N}(z_t; \sqrt{\alpha_t}z_0, (1 - \alpha_t)I).
\end{equation}

To reverse the diffusion process, a noise estimator $\epsilon_\theta(\cdot)$ parameterized by an UNet is learned to predict the forward added noise $\epsilon$ with the objective function,
\begin{equation}
\mathcal{L}_\text{dm} = \mathbb{E}_{(z_0,c)\sim D} \mathbb{E} _{\epsilon\sim\mathcal{N}(0,1),t}
\left[ ||\epsilon - \epsilon_\theta (z_t, t, c)||^2 \right],
\end{equation}
where $c$ is the text condition associated with image latent $z$, and $D$ is the training set. 

In MMTryon, our condition is not merely a text condition but a multi-modal condition that includes texture information as well as a multi-modal prompt that integrates global image information and instructions. Modulating this condition into the diffusion model is a challenging problem. Prior works~\cite{yang2023paint,chen2023anydoor} have relied on transforming pixel features into text-aligned features using a CLIP encoder and injecting them into the cross-attention layers. However, this approach often results in a loss of attention to detail. In this work, we instead draw inspiration from~\cite{hertz2022prompt,ge2023expressive,cao2023masactrl}, who show that the cross-attention layer learns correspondences between the image features and the text embedding tokens, generating an overall image semantic structure, while the self-attention layers are more related to detailed texture generation. We thus focus on enhancing the cross-attention layers conditioned on the original text, integrating it with our multi-modal instruction to improve generation of the final try-on results. Additionally, we train a garment encoder to align pixel features with the Stable Diffusion features, enhancing the self-attention layers and seamlessly transfer detailed image features from the garment images to the desired human images.

\begin{figure*}
  \centering
  \includegraphics[width=1.0\textwidth]{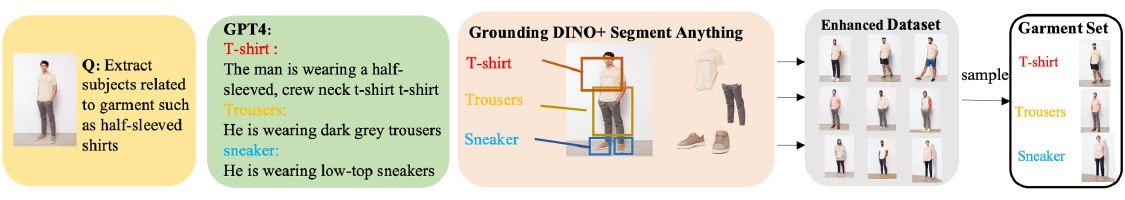}
\caption{The data generation pipeline of MMTryon. We use a large multi-modal model to describe the target person image, followed by open-vocabulary grounding and segmentation models to extract correspondences between a person image and several garment subjects. For each subject, we utilize SDXL inpainting to obtain the enhanced dataset, which serves as our training data. }
  \label{fig:datapipeline}

\end{figure*}

\subsection{Pretrain Garment Encoder}
As mentioned before, one of the main challenges of multi-reference try-on is the lack of training data, i.e., most of the garment person pairs consist of only a single garment. We draw inspiration from the training scheme of Large Multi-modal Models (e.g.,~\cite{liu2024visual}), and split the training into pretraining and supervised fine-tuning stages. During pretraining, our goal is to align features from multi-source to the generator, here, the Stable Diffusion denoising UNet. Thus we fixed the denoising UNet, and pretrained the garment encoder in this stage.

To achieve texture-consistent try-on results without relying on a prior segmentation model, this encoder should extract fine-grained image features and respond effectively to text queries.  We initialized the garment encoder with the UNet encoder from the diffusion model, which takes both garment image and text query as input, and use the features after the cross-attention layer of each transformer block as the output garment feature. This feature is then injected into the diffusion UNet through multi-reference attention. By supervising with diffusion loss $\mathcal{L}_\text{dm}$, we are able to obtain features that can preserve texture consistency.

Meanwhile, to improve the accuracy of text queries, we enforce the text-queried features to have a similar response to a prior segmentation model SAM~\cite{kirillov2023segment} (see Fig.~\ref{fig:pretrain}). Here we also introduce a text query loss, where features outside the region of the mask are passed through a sigmoid, squared, and averaged to encourage minimization of activations in areas unrelated to the text.

The final loss is: 
\begin{equation}
\mathcal{L}_\text{enc} = \mathcal{L}_{\text{dm}} + \mathcal{L}_{\text{query}}
\end{equation}
where the text query loss \(\mathcal{L}_{\text{query}}\) is:

\begin{equation}
\mathcal{L}_{\text{query}} = \frac{1}{N} \sum \sigma(F \odot (1 - M)) ^2
\end{equation}

Here, \(\odot\) denotes element-wise multiplication and \(\sigma\) denotes the sigmoid function. \(F\) represents the garment feature and \(M\) denotes the prior mask corresponding to the target region’s text. \(N\) denotes the number of elements in \(F\) and \(M\). This way, we can obtain a garment encoder that can extract only the necessary features corresponding to the text query which also preserves low-level pixel consistency through a combination of discriminative and generative supervision.

It is crucial to note that the reliance on prior masks from SAM is strictly confined to this pretraining phase. By optimizing $\mathcal{L}_{\text{query}}$, we effectively distill the segmentation capability into the garment encoder itself. we observe that SAM provides generally high-fidelity masks  offering reliable spatial supervision. While minor segmentation noise may exist, our cross-modal objective encourages the encoder to learn robust semantic features rather than overfitting to local boundary inconsistencies.This equips the encoder with an "emergent" ability to localize relevant garment features based solely on the textual label $P_i$. Consequently, during the inference phase, the model operates in a completely parsing-free manner, requiring no external segmentation masks or bounding boxes.

\subsection{MMTryon}

Here, we provide a detailed description of each component within the main UNet. In particular, we employ two types of attention mechanisms to enable multi-modal multi-reference try-on. Having established the powerful garment encoder in the previous section, we now eliminate the dependency on prior masks by utilizing a synthetic dataset to finetune the entire model and the garment encoder jointly. This dataset is derived from the original data through an enhanced data generation pipeline, which is detailed in Sec.~\ref{sec:data}.

\subsubsection{Scalable Data Generation Pipeline}
\label{sec:data}
To achieve multi-modal multi-reference try-on, a straightforward way is to construct training data in the same format. However, existing try-on datasets in this format are not readily available. There are generally two types of datasets, in-shop garment to person pairs or person to person pairs, where the former one only contains a single reference and is usually limited in garment types. 

Leveraging the person-to-person pairs instead, a possible way to automatically construct a training dataset for the multi-modal multi-reference try-on task is to use segmentation models to crop various reference garments to construct multi-reference pairs. However, the drawback of this approach is that the style of the garments is not controllable and that the model will still be dependent on off-the-shelf segmentation models during inference.

In this work, we instead propose to leverage the increasingly powerful capabilities of large models to develop a flexible scheme for dataset construction. Starting from existing person-to-person datasets, given a pair of person images $I_a, I_b$, we use SOTA large multi-modal model GPT-4V~\cite{achiam2023gpt} and CogVLM~\cite{wang2023cogvlm} to caption the image in a desired format describing how the person is dressed. We then extract the main garment subjects and style (garment categories) from the image description denoted as $\langle obj_1, obj_2, \ldots, obj_n \rangle$. Subsequently, we use a SOTA open-vocabulary detection model~\cite{liu2023grounding} to generate the bounding boxes corresponding to each garment subject within the image. The SAM model~\cite{kirillov2023segment} is then used to generate precise mask annotations. After this caption-grounding-segmentation procedure, we now establish a correspondence between a target person image $I_a$ with description, and all its segmented garments from $I_b$. We find that this process performs well for the vast majority of clothing categories. Note that we need to establish correspondences between two different poses, otherwise this try-on task will degenerate to a copy-paste problem.

To enable parsing-free inference, starting from those garment images, we then use the inpainting ability of the Stable Diffusion model (we use SDXL~\cite{podell2023sdxl} here) with a given prompt to randomly generate an image where a person is wearing such a garment. We use random seeds and random prompts to generate multiple samples, and discard samples with high similarity to prevent information leakage. This way, by iterating through all garment subjects, we are able to construct training pairs, where the target image is described using text, and the reference garments are from different images. The overall process is illustrated in Fig.~\ref{fig:datapipeline}.

\subsubsection{Multi-Modal Instruction Attention}
In our task, it is not possible to ask the user to describe the desired garments from the reference images precisely using only text, as high-level descriptions such as garment type exhibit ambiguity and cannot guarantee dressing fidelity. To mitigate this issue, we form a multi-modal instruction using a prompt template: 

\begin{small}
    \[
    \text{A person wearing  \textless garment\textgreater \textless style\textgreater }
    [Ref\#1] \\
    \text{..., }[Ref\#N]
    \]
\end{small}

For example, {\em A person wearing a top, tucked in, [Ref\#1], pants [Ref\#2] and shoes [Ref\#3]}. Here {\em [Ref\#N]} is a special placeholder token prepared for the corresponding reference image. This text prompt is first encoded using CLIP text encoders into a text embedding $\mathbf{T}\in\mathbb{R}^{L\times D}$, where $L$ is the length of the tokens and $D$ is the embedding dimension. The embeddings corresponding to the placeholder tokens are then fused with the CLIP image embedding of the related garment images (see Fig.~\ref{fig:framework}). To balance the embedding length and image feature granularity, and also better align the image feature with the text embedding, we do not only use the [CLS] token as in~\cite{yang2023paint} 
, but instead use the latent features from the penultimate layers of the CLIP ViT encoder. These latent features are further downsampled using a perceiver-resampler~\cite{alayrac2022flamingo}.
Specifically, when handling multiple garment reference images, we extract their corresponding CLIP visual features to serve as unique identifiers, effectively binding the specific garment characteristics with the textual context. Consequently, the final instruction becomes a seamless fusion of garment CLIP features and text features. This fused multi-modal embedding replaces the original text embedding of the Stable Diffusion model and is injected into the cross-attention layers of the UNet. By facilitating direct interaction between the visual cues and textual tokens.
This multi-modal embedding then serves as the replacement of the original text embedding of the Stable Diffusion model and not only aids in the generation of dressing styles but also enhances the stability of the try-on results, ensuring the correct mapping of the garment. 

\subsubsection{Multi-Reference Texture Attention}

The Multi-Modal Instruction Attention cannot fully preserve the details of the garment due to the use of the CLIP ViT image encoder. To guarantee texture consistency, we therefore first prompt the garment encoder with the textual label. The textual label $P_i$ is derived from the instruction prompt $T$ and used to extract features for the particular subject from image $G_i$.
After obtaining the features, in each attention layer of the diffusion UNet, we extract a set of garment features corresponding to that layer, denoted as \(\{F_{G_1}, F_{G_2}, \ldots, F_{G_n}\}\). We denote the original feature map from the diffusion UNet as \(F_\text{target}\).
To seamlessly transfer garment textures while preserving the target person's structure, we \textbf{rethink the role of the standard self-attention mechanism}. Instead of limiting the interaction to the target image itself, we aim to establish a \textbf{unified pixel-level interaction space} where the target feature can simultaneously attend to the reference garments and its own internal structure. To achieve this, we construct a \textbf{hybrid attention mechanism}: the Query ($Q$) is derived from \(F_\text{target}\), while the Key ($K$) and Value ($V$) are obtained from the concatenation of the target and garment features:
\begin{equation}
    K, V = \text{Concat}([F_\text{target}, F_{G_1}, \ldots, F_{G_n}])
\end{equation}
Crucially, the rationale behind including \(F_\text{target}\) in the Key and Value pairs is two-fold:
\begin{itemize}
    \item \textbf{Structure Preservation (Self-Attention):} By allowing the query to attend to the \(F_\text{target}\) component within \(K\), the model acts as a standard self-attention layer. This ensures that the generated image maintains the structural integrity, body pose, and identity of the non-editing regions (e.g., face, background) by attending to its own spatial features.
    \item \textbf{Implicit Warping (Cross-Attention):} By allowing the query to attend to the \(\{F_{G_i}\}\) components within \(K\), the model performs a dense texture lookup. This acts as an \textbf{implicit warping} mechanism, where the target pixels semantically align with and absorb fine-grained texture details from the reference garments without requiring an explicit warping module.
\end{itemize}
This hybrid multi-reference attention allows us to harmonize the generation process, effectively preserving the texture features of each object via implicit warping while maintaining the global structure via self-attention. The process is illustrated in Fig.~\ref{fig:framework}.

\section{Experiment}

\subsection{Datasets}

We conduct extensive experiments on two public high-resolution Virtual Try-ON (VITON) benchmarks, namely VITON-HD~\cite{choi2021viton} and DressCode~\cite{davide2022dress}, as well as on an additional large-scale proprietary e-commerce dataset. Experiments are conducted under a resolution of $1024 \times 768$. Specifically, VITON-HD comprises 13,679 image pairs of front-view upper-body women and upper-body in-store garment images, which are further divided into 11,647 training pairs and 2,032 testing pairs. DressCode includes 48,392 training pairs and 5,400 testing pairs of front-view full-body person and in-store garment images, consisting of three subsets with different category pairs (i.e., upper, lower, dresses).
Our proprietary e-commerce dataset, which will be released upon acceptance, contains 203,239 training pairs and 5,219 testing pairs of front-view full-body person and in-store garment images, and 450,931 training pairs and 9,198 testing pairs of multiple front-view images of the same full body.

\begin{figure}
  \centering
  \includegraphics{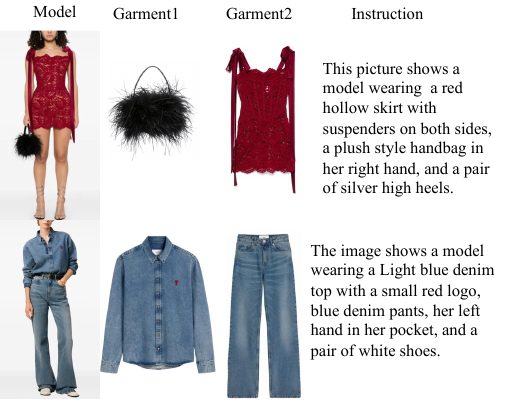}
  \caption{Examples of our multi-reference multi-modal instruction dataset.}
  \label{fig:data_vis}
\end{figure}

\subsection{Implementation Details} \label{sec:implemet_details}

During the data augmentation process, we enhanced the data across the three datasets, running the entire enhancement workflow on 16 A800 GPUs for 5 days. The augmented dataset comprises approximately 2 million images. In our MMTryon implementation, we utilize the SD 1.5 pretrained model~\cite{rombach2022high} to initialize the weights of our main UNet and the weights of our pretrained garment encoder. For the encoding of both text and global image information, we use clip-large-14~\cite{radford2021learning}. The garment encoder was pretrained on 8 A100 GPUs for 3 days with a total batch size of 8 and MMTryon was trained on 8 A100 GPUs for 1 day with a total batch size of 8.

\subsection{Metric}
\begin{figure}
  \centering
  \includegraphics[width=0.5\textwidth]{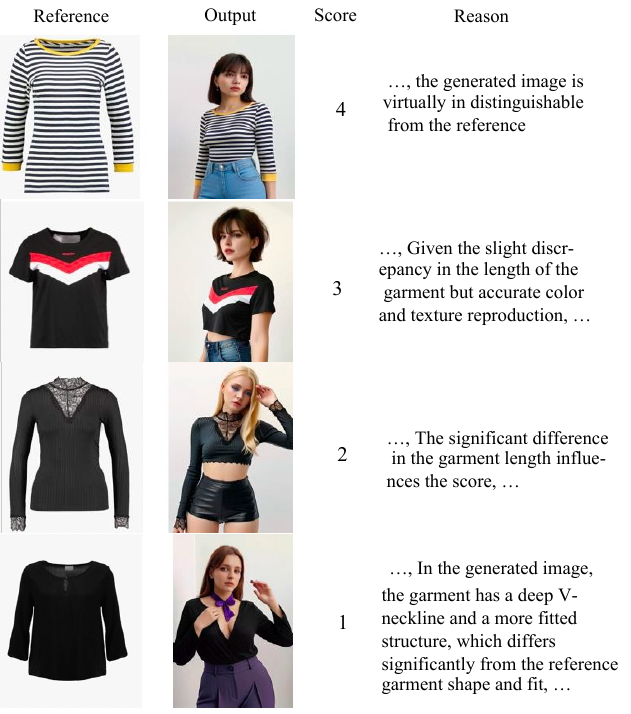}
  \caption{Visualized results of the outputs and evaluation reasons corresponding to different scores.}
  \label{fig:HAS_visulize}
\end{figure}

\begin{figure*}

  \centering
  \includegraphics{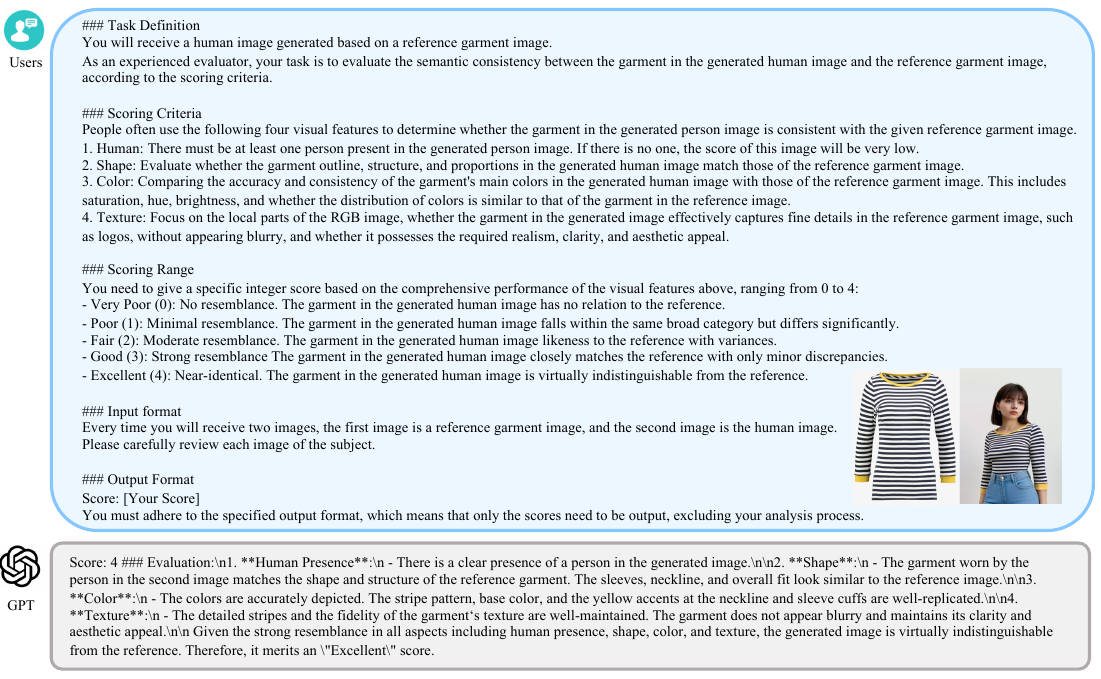}
  \caption{The evaluation pipeline to obtain the Human-Aligned Score.} 
  \label{fig:HAS_pipe}

\end{figure*}

For traditional single-garment try-on tasks, we utilize four widely-used metrics, namely the Structural Similarity Index (SSIM)~\cite{wang2004image}, Perceptual Distance (LPIPS)~\cite{zhang2018unreasonable}, Kernel Inception Distance ~\cite{sutherland2018demystifying}, and Fréchet Inception Distance (FID)~\cite{parmar2022aliased}, to evaluate the similarity between synthesized and real images. 
For the compositional try-on task, we similarly employ FID and KID to measure the image generation quality. In the context of the multi-modal instruction try-on task, we assess the image generation quality using FID, while the Clip score is utilized to evaluate the consistency between instructions and the final generated images. 
Additionally, we conducted a Human Evaluation (HE) study, inviting 100 participants to assess the synthesis quality and text-image consistency across single-garment try-on scenarios.
We further employ GPT-4o~\cite{achiam2023gpt} for automatic evaluation of human preference, denoted as Human-Aligned Score (HAS). 
To design the evaluation prompts, we drew inspiration from DreamBench++~\cite{peng2024dreambench++}, making adjustments to the task definitions, scoring criteria, and scoring range to better suit the Garment-Centric Human Generation task.
Fig.~\ref{fig:HAS_visulize} showcases output examples along with corresponding evaluation rationales for different scores. GPT-4o assigns scores on a scale from 0 to 4, which are then normalized to a range of 0 to 1 for quantitative evaluation purposes. Fig.~\ref{fig:HAS_pipe} shows the detail of the evaluation pipeline.

\subsection{Baselines}
\label{sec:baselines}

\begin{table*}[t]
\setlength{\tabcolsep}{2.6mm}
\centering
\caption{The sample of Tag generation questions. }
\begin{tabular}{|p{1\textwidth}|}
  \toprule
  \textbf{Question} \\
  \cmidrule(lr){1-1}
  Is this image a garment or a model? Use one word. \\
  \cmidrule(lr){1-1}
  Describe this picture in detail. \\
  \cmidrule(lr){1-1}
  What is the sleeve length of this garment? Is it long sleeves, short sleeves, or sleeveless?  \\
  \cmidrule(lr){1-1}
  If it's long sleeves, are the sleeves rolled up? \\
  \cmidrule(lr){1-1}
  How long are the sleeves? Do they reach the wrist, elbow, or shoulder? . \\
  \cmidrule(lr){1-1}
  Does this garment have a visible logo or brand marking?  \\
  \cmidrule(lr){1-1}
  What is the neckline style? Is it a round neck, V-neck, or another type? \\
  \bottomrule
\end{tabular}
\label{tab:questions}
\end{table*}

In our study, we initially compare our approach on traditional try-on tasks with previous baseline methods. We selected four advanced GAN-based methods, namely PF-AFN~\cite{ge2021parser}, FS-VTON~\cite{he2022style}, SDAFN~\cite{bai2022single}, GP-VTON~\cite{xie2023gp} and the latest diffusion-based methods, DCI-VTON~\cite{gou2023taming} and StableVTON \cite{kim2023stableviton}. We directly utilize their released pretrained models for this comparison. Unfortunately, we are not able to conduct an extensive comparison with 
TryonDiffusion~\cite{zhu2023tryondiffusion} as it is not publicly available at the time of submission. 
Secondly, for the compositional and multi-modal try-on tasks, we compare our work with paint by example~\cite{yang2023paint}, Stable Diffusion~\cite{rombach2022high}, and DALLE3~\cite{dalle3}. 
For all baselines, we strictly adhere to the official instructions for running their training and testing scripts.

\subsection{Training Dataset Details}\label{sec:TD}

The dataset was sourced from authorized e-commerce websites and we have obtained the necessary permissions for the use of the data and the models featured within it. 

To mitigate inherent biases and domain gaps, we utilized diverse generation models during data creation and enforced a strict human-in-the-loop filtering process.
Specifically, we employed seven distinct types of prompts to guide the SDXL model for inpainting, where each type corresponds to a specific garment category (e.g., trousers, skirts) and encompasses a wide range of dressing styles. 
Beyond SDXL-based inpainting, we further enriched our dataset complexity by leveraging established high-quality single-reference try-on models to generate diverse data samples. 
All generated data underwent rigorous manual screening to ensure the highest quality. We strictly retained only samples that exhibited high photorealism and anatomically correct human structures, discarding any images with artifacts or distortions.

\textbf{Garment-to-Person and Synthetic Data Generation:} To further enhance the dataset, we trained a single-reference try-on network. Using this model, we transformed the garment-to-person data into person-to-person pairs, generating synthetic garment-person-person combinations. This resulted in a multi-reference multi-modal instruction dataset that combines synthetic multi-reference data with single-reference garment-person pairs.
Fig~\ref{fig:data_vis} shows some examples from our dataset.

\textbf{Ensuring Dataset Diversity and Quality:} To ensure the dataset is representative of various demographics, we developed a comprehensive labeling system that accounts for attributes such as gender, skin tone, age, and body type. This labeling system aims to ensure that the model performs well across a wide range of users. Additionally, we utilized \textbf{CogVLM}~\cite{wang2023cogvlm} to annotate and balance the dataset based on various attributes, including garment categories, styles, and other relevant features. Example prompts for this step are provided in Table~\ref{tab:questions}. These garment-person pairs encompass a wide variety of body types and characteristics.

\textbf{Data Release and Reproducibility:} To facilitate future research, we are committed to releasing the full configuration of our augmentation pipeline. Furthermore, we will provide a high-quality augmented subset of approximately 500,000 images derived from public open-source datasets to the research community.


\subsection{Human Evaluation Details}\label{sec:HE}
For the human evaluation, we designed questionnaires for the single garment try-on, multi-modal garment try-on, and compositional garment try-on. 100 volunteers were invited to complete a questionnaire comprised of 50 assignments for each task. Specifically, for the single garment try-on, given a person image and a garment image, volunteers were asked to select the most realistic and accurate try-on result out of nine  options generated by our MMTryon and the baseline methods (i.e., PF-AFN~\cite{ge2021parser}, FS-VTON~\cite{he2022style}, SDAFN~\cite{bai2022single}, GP-VTON~\cite{xie2023gp}, DCI-VTON~\cite{gou2023taming}, Stable-VTON~\cite{kim2023stableviton}), anydoor \cite{chen2023anydoor} and Ootdiffusion \cite{xu2024ootdiffusion}. 

\begin{figure*}
  \centering
  \includegraphics[width=1.0\textwidth]{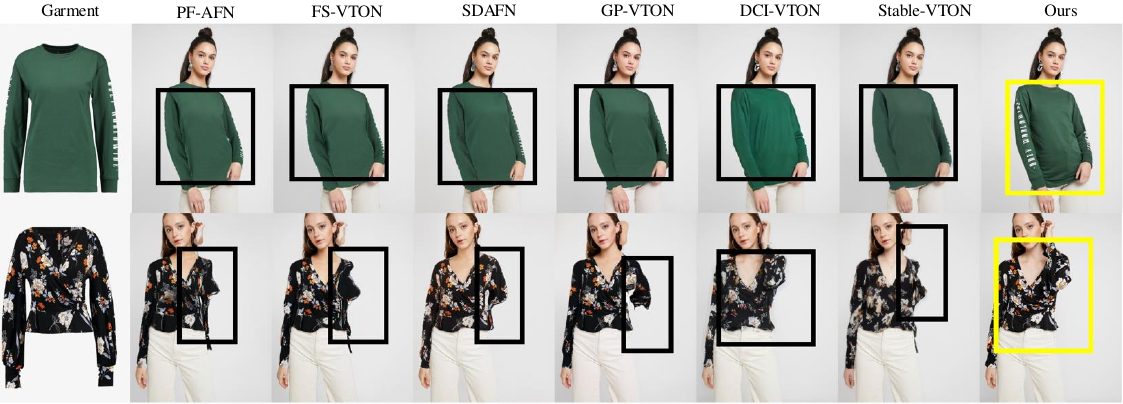}

  \caption{Qualitative comparisons on VITON-HD in the single garment try-on task. MMTryon produces more realistic and texture-consistent images. 
  }
  \label{fig:single}
\end{figure*}

\begin{figure}
  \centering
  \includegraphics[width=0.5\textwidth]{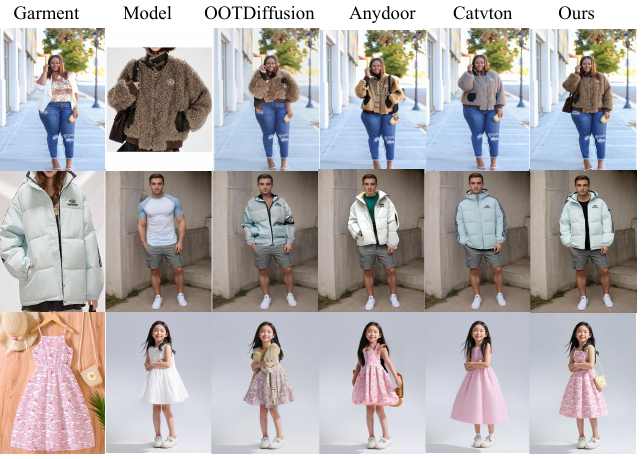}
  \caption{Qualitative comparisons for in-the-wild single-garment try-on.}
  \label{fig:wild}
\end{figure}

The order of the generated results in each assignment was randomly shuffled.
For the compositional garment try-on, given a person image and a set of garment images, volunteers were asked to select the most realistic and accurate try-on result out of four options generated by our MMTryon and the baseline methods (i.e., Paint by example~\cite{yang2023paint}, Midjourney, DALLE3~\cite{dalle3}, Nano Banana~\cite{gemini2023}) .
For the multi-modal garment try-on, given a person image, a set of garment images, and a text instruction, volunteers were asked to select the most realistic and instructionally relevant try-on result out of four options generated by our MMTryon and baseline methods (i.e., Paint by example~\cite{yang2023paint}, Midjourney, DALLE3~\cite{dalle3},Nano Banana~\cite{gemini2023}).

\subsection{Qualitative Results}

\noindent \textbf{Comparison on single garment try-on.} 
Fig.~\ref{fig:single} provides qualitative comparisons between MMTryon and the state-of-the-art baselines in the single garment try-on task on the VITON-HD dataset~\cite{choi2021viton}. The results demonstrate the superiority of our MMTryon over the baselines. Firstly, our method employs a more powerful garment encoder, which results in better texture consistency compared to the baselines. Secondly, as our approach utilizes more textual information, it achieves greater stability in try-on styles than the baseline, showcasing the effectiveness of our method. Furthermore, we have updated Fig.~\ref{fig:wild} to include evaluations on challenging in-the-wild scenarios, which more explicitly highlights the robustness and superior generalization capabilities of our approach.

\begin{figure*}

  \centering
  \includegraphics[width=1.0\textwidth]{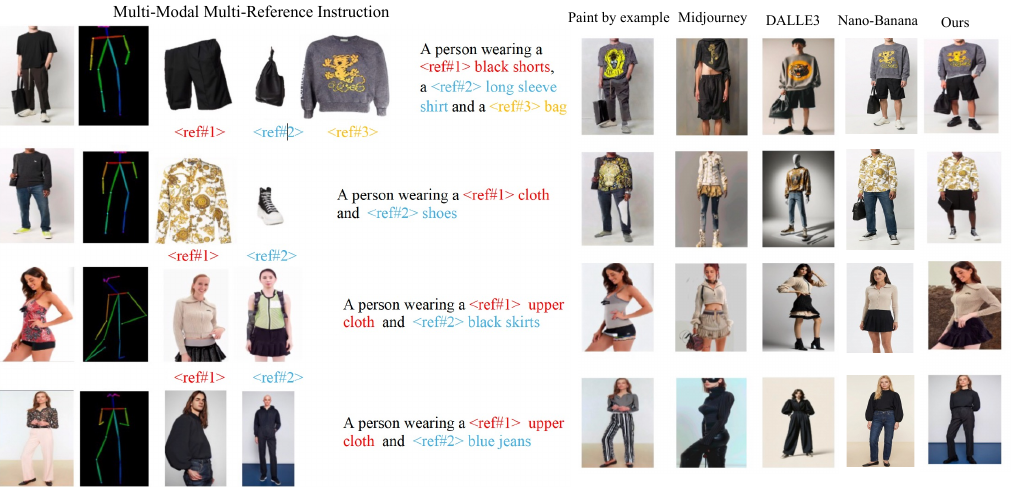}
  \caption{Qualitative comparisons with Paint-by-Example~\cite{yang2023paint}, Midjourney  DALLE3~\cite{dalle3} and Nano Banana~\cite{gemini2023} in the Multi-Modal Multi-Reference try-on task. Compared with other methods, our method can achieve the best results both in terms of faithfulness and realism.} 
  \label{fig:Multi}
\end{figure*}

\begin{figure*}[t]
  \centering
  \includegraphics[width=1.0\textwidth]{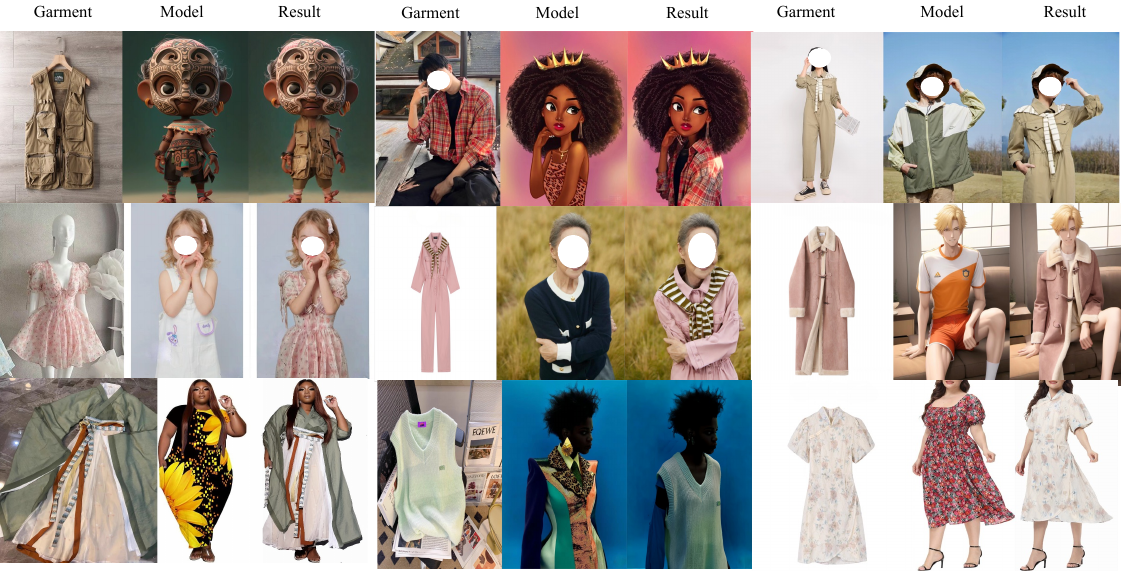}
  \caption{Results of our MMTryon, MMTryon demonstrates superior performance in cartoon character, in-the-wild, and diverse virtual try-on scenarios} 

\label{fig:supp3}
\end{figure*}

\noindent \textbf{Comparison on multi-modal and compositional try-on.} Fig.~\ref{fig:Multi} provides a qualitative comparison between MMTryon and the state-of-the-art baselines for the compositional try-on and multi-modal try-on tasks in the wild. The results demonstrate the superiority of our approach relative to the baselines. Firstly, for compositional try-on, our method exhibits better texture consistency and more realistic try-on effects compared to the baselines. 
Secondly, for multi-modal try-on, our approach demonstrates more refined performance in terms of texture consistency.
At the same time, baselines often overlook the textual descriptions of dressing styles, whereas our method demonstrates stronger text-image consistency, effectively reflecting the intentions expressed in the text. This demonstrates that our approach achieves state-of-the-art results for the compositional and multi-modal try-on tasks. Additional qualitative results of our method are included in Fig. ~\ref{fig:supp3}. 

\begin{table}[t]
\centering
\small
\setlength{\tabcolsep}{1.5pt}
\caption{Quantitative comparisons on the single garment try-on task. We report SSIM ($\uparrow$), FID ($\downarrow$), LPIPS ($\downarrow$), KID ($\downarrow$), and Human Evaluation (HE $\uparrow$). ``-'' indicates the metric is not available.}
\begin{tabular}{l c c c c c}
  \toprule
  Method & SSIM $\uparrow$ & FID $\downarrow$ & LPIPS $\downarrow$ & KID $\downarrow$ & HE $\uparrow$ \\
  \midrule
  PF-AFN~\cite{ge2021parser}      & 0.885 & 9.616 & 0.087 & 3.85 & 0.03 \\
  FS-VTON~\cite{he2022style}      & 0.881 & 9.735 & 0.091 & 3.69 & 0.02 \\
  SDAFN~\cite{bai2022single}      & 0.887 & 9.497 & 0.092 & 2.73 & 0.03 \\
  GP-VTON~\cite{xie2023gp}        & 0.893 & 9.405 & 0.079 & 0.88 & 0.10 \\
  DCI-VTON~\cite{gou2023taming}   & 0.868 & 9.166 & 0.096 & 1.10 & 0.10 \\
  StableVTON~\cite{kim2023stableviton} & 0.866 & 8.992 & 0.079 & 1.03 & 0.16 \\
  OOTDiffusion~\cite{xu2024ootdiffusion} & 0.878 & 8.810 & 0.071 & 0.82 & 0.10 \\
  AnyDoor~\cite{chen2023anydoor}  & 0.862 & 12.735& 0.149 & 5.87 & 0.06 \\      
  Wear-Any-Way~\cite{chen2024wear}&0.877 & 8.155 & 0.078 & 0.78 & - \\
  AnyFit~\cite{li2024anyfit}      &  0.881 & 8.600 & 0.075 & 0.55 & - \\
  CatVTON~\cite{chong2025catvton}      & 0.870 & \textbf{8.010}  & 0.071 & \textbf{0.51} & 0.18 \\
  \midrule
  \textbf{MMTryon (Ours)}         & \textbf{0.902} & 8.702 & \textbf{0.069} & 0.58 & \textbf{0.22} \\  
  \bottomrule
\end{tabular}
\label{tab:single}
\end{table}

\subsection{Quantitative Results}

As reported in Tab.~\ref{tab:single}, Tab.~\ref{tab:compositional} and Tab.~\ref{tab:multimodal}, While MMTryon achieves comparable performance to SOTA methods on distribution-based metrics (e.g., FID/KID), it significantly outperforms them in structure preservation (SSIM) and human preference (HE), proving that MMTryon can generate try-on results with better visual quality in single garment try-on, compositional try-on and multi-modal try-on. Furthermore, in our human evaluation, we compare the generated results of the baselines to our model. Higher human evaluation scores indicate that a larger proportion of participants prefer the outcomes of the given method. The human evaluators prefer the results generated by MMTryon in most cases, indicating superior texture consistency and text-image consistency.

\subsection{Ablation study}
\begin{figure*}

  \centering
\includegraphics[width=1.0\textwidth]{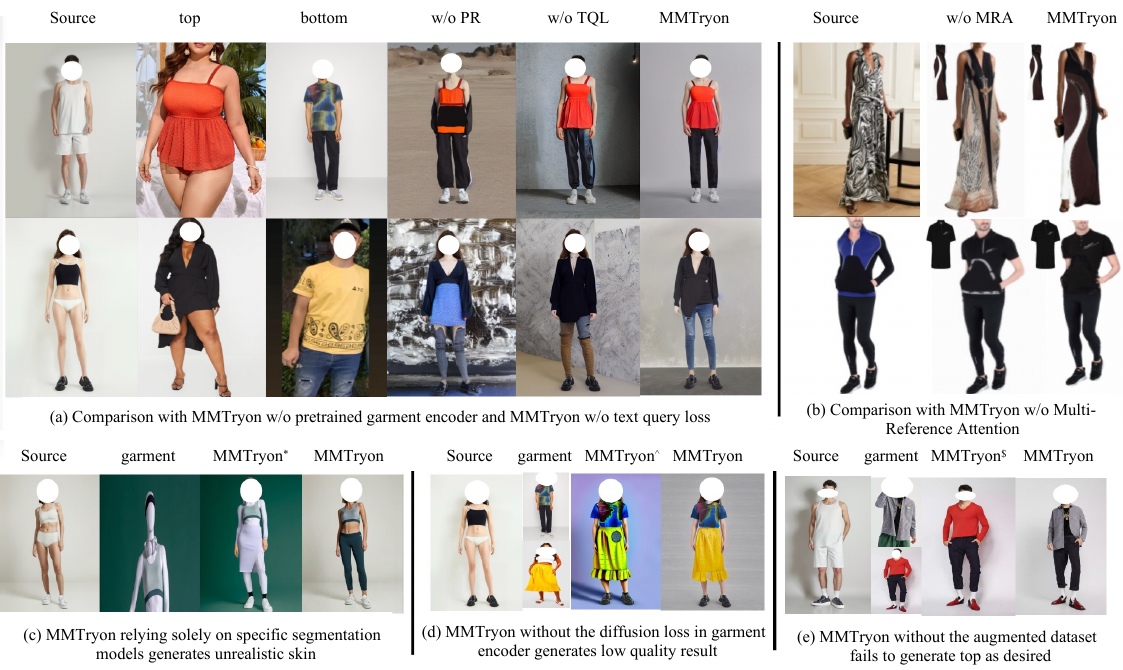}

    \caption{
    Qualitative results of the ablation study. Here, MMTryon$^*$ denotes the variant relying solely on specific segmentation models. MMTryon$^\wedge$ represents the model trained without the diffusion loss in the garment encoder, and MMTryon$^\$$ refers to the variant trained without our augmented dataset.}
    \vspace{2mm}
  \label{fig:ablation}

\end{figure*}
\textbf{Effectiveness of the pretrained garment encoder}:
To validate the effectiveness of the pretrained garment
encoder, we conduct an ablation study by comparing MMTryon with two ablated versions. One that directly trains the entire model without pretrained garment encoder (denoted as w/o PR) and one that does not include the text query loss (denoted as w/o TQL). The results, as shown in Fig.~\ref{fig:ablation}(a)
, indicate that omitting the pretrained garment encoder or text query loss will lead to a decrease in garment consistency and overall generation quality. The quantitative results, as presented in Tab.~\ref{tab:ablation} 
demonstrate that MMTryon with pretrained garment encoder and supervised by the text query loss significantly outperforms the other two versions as the garment consistency score and overall generation quality score are both higher. 

\begin{table}[t]
\centering
\setlength{\tabcolsep}{1.5pt}
\caption{Quantitative comparisons on the compositional garment try-on task.(Compositional Try-on)}
\small
\begin{tabular}{l c c c c c c}
  \toprule
  Method & SSIM $\uparrow$ & FID $\downarrow$ & LPIPS $\downarrow$ & KID $\downarrow$ &HE $\uparrow$&\\
  \midrule
  Paint-by-Example & 0.423 & 12.429 & 0.182 & 4.89 & 0.02  \\
  Midjourney & 0.699 & 9.231& 0.083  & 3.41 & 0.13 \\
  DALL·E3 & 0.763 & 9.010 & 0.079  & 1.03 & 0.23 \\
  Nano Banana & 0.798 & 8.926 & 0.081 & 0.72 & 0.26 \\
    \midrule
  \textbf{MMTryon (ours)}& \textbf{0.872}  & \textbf{8.902} & \textbf{0.073} & \textbf{0.47} & \textbf{0.36} \\
  \bottomrule
\end{tabular}
\label{tab:compositional}
\end{table}



\begin{table}[t]
\centering

\caption{Quantitative comparisons on the compositional garment try-on task.(Multi-Modal Try-on)}
\small
\tabcolsep 4.0 pt
\begin{tabular}{l c c c c c}
  \toprule
  Method & FID $\downarrow$ & KID $\downarrow$ & CLIP $\uparrow$ & HAS $\uparrow$ & HE $\uparrow$ \\
  \midrule
  Paint-by-Example & 15.231 & 5.21 & 0.06 & 0.41& 0.01 \\
  Midjourney & 10.243 & 3.45 & 0.08 & 0.46& 0.09 \\
  DALL·E3 & 9.132 & 1.39 & 0.19 &0.76& 0.28 \\
  Nano Banana & 8.343 & 0.99 & 0.22 &0.88& 0.24 \\
  \midrule
  \textbf{MMTryon (ours)} & \textbf{8.187} & \textbf{0.42} & \textbf{0.30} & \textbf{0.93} & \textbf{0.38} \\
  \bottomrule
\end{tabular}
\label{tab:multimodal}

\end{table}

\begin{table}

\centering
\caption{Ablation result demonstrate the benefit of the pretrained garment encoder (PR), text query loss (TQL), and multi-reference attention (MRA).}

\small
\tabcolsep 12pt
\begin{tabular}{l c c c c c}
  \toprule
  Method & FID $\downarrow$ & KID $\downarrow$ & CLIP Score $\uparrow$ \\
 \cmidrule(lr){1-1}\cmidrule(lr){2-4}
  w/o PR & 10.230 & 0.68 & 0.22 \\
  w/o TQL & 9.250 & 0.65  & 0.24 \\
  w/o MRA & 12.890    & 0.87   & 0.18\\
  \cmidrule(lr){1-1}\cmidrule(lr){2-4}
  \textbf{MMTryon} & \textbf{8.187} & \textbf{0.42}  & \textbf{0.30} \\
  \bottomrule
\end{tabular}
\label{tab:ablation}

\end{table}

\textbf{Effectiveness of the multi-reference attention}:
To further verify the effectiveness of the multi-reference attention, we compare MMTryon with a version that only utilizes the CLIP features of the clothing, denoted as w/o MRA, where we replaced the multi-reference attention with the original self-attention module. 
We again compare the main metric scores and observe a decline in both image quality and texture consistency when multi-reference attention is removed. Meanwhile, Fig.~\ref{fig:ablation}(b) indicates that rich clothing features are key to maintaining texture consistency. 

\textbf{Effectiveness of the large-scale segmentation model:}
To validate the necessity of integrating large-scale, open-vocabulary models in instruction-based try-on, we conducted further ablation studies. While domain-specific segmenters may work well in specific cases, relying on them introduces artifacts and errors. To mitigate these issues, we combine the Text-Query Loss and the Reconstruction Loss, allowing MMTryon to train end-to-end. Fig.~\ref{fig:ablation}(c) illustrates the limitations of traditional segmenters, which are prone to errors from incorrect annotations, highlighting the need for a more robust solution.

\textbf{Garment Encoder Without Diffusion Loss}:
We performed an additional ablation to evaluate the role of the diffusion loss on the garment encoder’s performance. When the diffusion loss was removed, the encoder failed to converge properly. This can be attributed to the lack of alignment with the main UNet, which the diffusion loss helps to maintain. In our experiments, we used only the Text-Query Loss to train the garment encoder. However, as shown in Fig.~\ref{fig:ablation}(d), the results were suboptimal, underscoring the importance of the diffusion loss in achieving accurate garment encoding.

\textbf{Effectiveness of Augmented Dataset:}
To further validate the contribution of our augmented dataset, we conducted an ablation study by training a variant of MMTryon (denoted as MMTryon\$) using the dataset without high-quality inpainting. Fig.~\ref{fig:ablation}(e) illustrates that this variant struggled with query accuracy, resulting in less precise try-on outcomes. In contrast, the full MMTryon model, trained with the complete augmented dataset, significantly outperformed its counterpart in understanding and responding to text-based queries. These findings highlight the crucial role of high-quality data augmentation in enhancing try-on performance.

\begin{figure}
  \centering
  \includegraphics[width=0.5\textwidth]{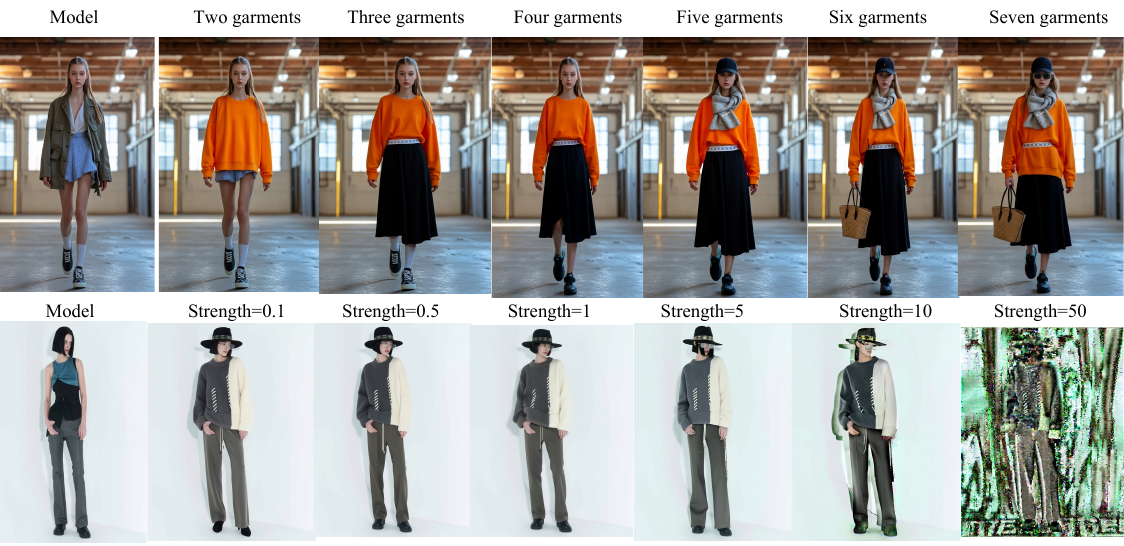}
  \caption{Ablation results of varying condition strengths.}
  \label{fig:ab_con}


\end{figure}

\textbf{Influence of Condition Strength}:
To justify our design choices, we evaluate the impact of condition strength across two dimensions. First, we investigate the influence of the number of garment reference images by testing inputs ranging from two to seven references. As illustrated in Fig.~\ref{fig:ab_con}, providing more reference images within this range offers richer texture cues and progressively enhances the texture quality of the generated results. However, exceeding seven references introduces conflicting information that compromises the structural correctness of the generated garments. Second, we analyze the effect of text conditioning by adjusting the guidance scale across several intensity levels. A moderate scale achieves the optimal balance between instruction adherence and natural image fidelity, whereas an excessively high conditioning strength results in blurry outputs and degrades the overall generation quality.

\subsection{Failure Case Analysis}

\begin{figure}
  \centering
  \includegraphics[width=0.5\textwidth]{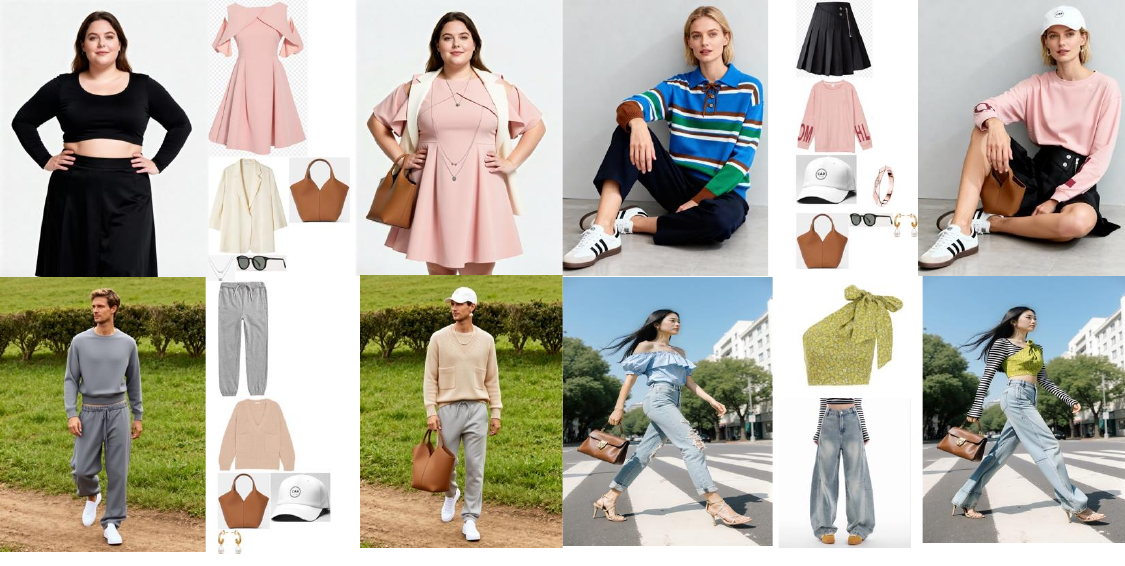}

  \caption{Visualized results of failure cases.}
  \label{fig:Fail}

\end{figure}

Despite the superior performance of MMTryon, certain limitations persist in complex scenarios, as illustrated in Fig.~\ref{fig:Fail}.First, the model occasionally generates unintended artifacts, such as the erroneous appearance of a necklace on a garment due to the denoising process's tendency to over-associate specific necklines with jewelry. Second, fine-grained category support remains challenging; for instance, small accessories like sunglasses may fail to be rendered in the final output. Third, the inherent stochasticity of the diffusion process can lead to unintended modifications in non-target regions, such as slight changes in footwear styles across different random seeds.  Fourth, while our multi-reference attention mechanism enables the model to scale to multiple garments—with six items being highly sufficient for most real-world scenarios—we observe a degradation in structural correctness when this number is significantly exceeded. Lastly, the model may encounter structural errors when processing out-of-distribution or "eccentric" clothing types that are underrepresented in the current training set. 
We attribute these issues primarily to the insufficient depth and breadth of the current dataset. We plan to continuously refine the data generation pipeline by incorporating more diverse garment categories, extreme multi-item compositions, and fine-grained annotations to further enhance the model's robustness and scalability.

\section{Conclusion and Discussion}\label{sec:conclusion}
\noindent\textbf{Conclusion.}
In this work, we introduce MMTryon, a novel and powerful try-on model capable of freely generating high-fidelity VITON results with realistic try-on effects based on text and multiple garments. By using a pretrained encoder, MMTryon not only avoids the need for segmentation but also achieves compositional try-on even with limited data. To support multi-modal and multi-reference dressing modes, MMTryon introduces multi-modal instruction attention and multi-reference attention modules. Experiments conducted on high-resolution VITON benchmarks and in-the-wild test sets demonstrate MMTryon's superior efficacy compared to existing methods.

\noindent\textbf{Limitation and Future Work.}
While our method demonstrates strong performance, it still has certain limitations. In the data generation process, our method is influenced by the limitations of pretrained models, making it challenging to produce data that meets the requirements for very fine parts, such as cuffs and collars. This restricts our ability to generate detailed components. Moving forward, we may focus on fine-tuning large models to construct a more freely detailed and fine-grained dataset, aiming to enhance the upper limit of our model.


 

\bibliography{zhangxujie}
\bibliographystyle{IEEEtran}

\begin{IEEEbiography}[{\includegraphics[width=1in,height=1.25in,clip,keepaspectratio]{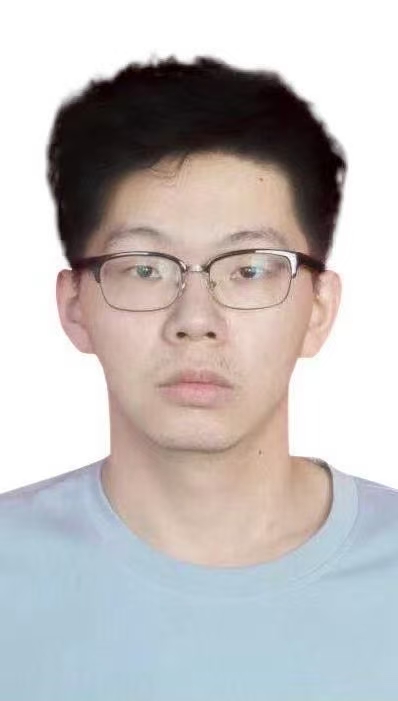}}]{Xujie Zhang} received his BE degree from Sun Yat-sen University, China. Now, he is working towards his PHD degree in Sun Yat-sen University . His research interests include Human 2D synthesis.
\end{IEEEbiography}

\begin{IEEEbiography}[{\includegraphics[width=1in,height=1.25in,clip,keepaspectratio]{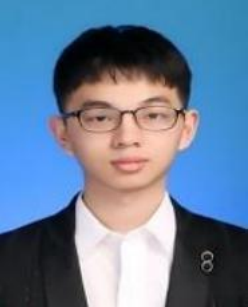}}]{Ente Lin} Ente Lin received his BE degree from Nanjing University of Aeronautics and Astronautics, China. Now, he is working towards his ME degree in Tsinghua University. His research interests include 2D synthesis and Embodied Intelligence.

\end{IEEEbiography}

\begin{IEEEbiography}[{\includegraphics[width=1in,height=1.25in,clip,keepaspectratio]{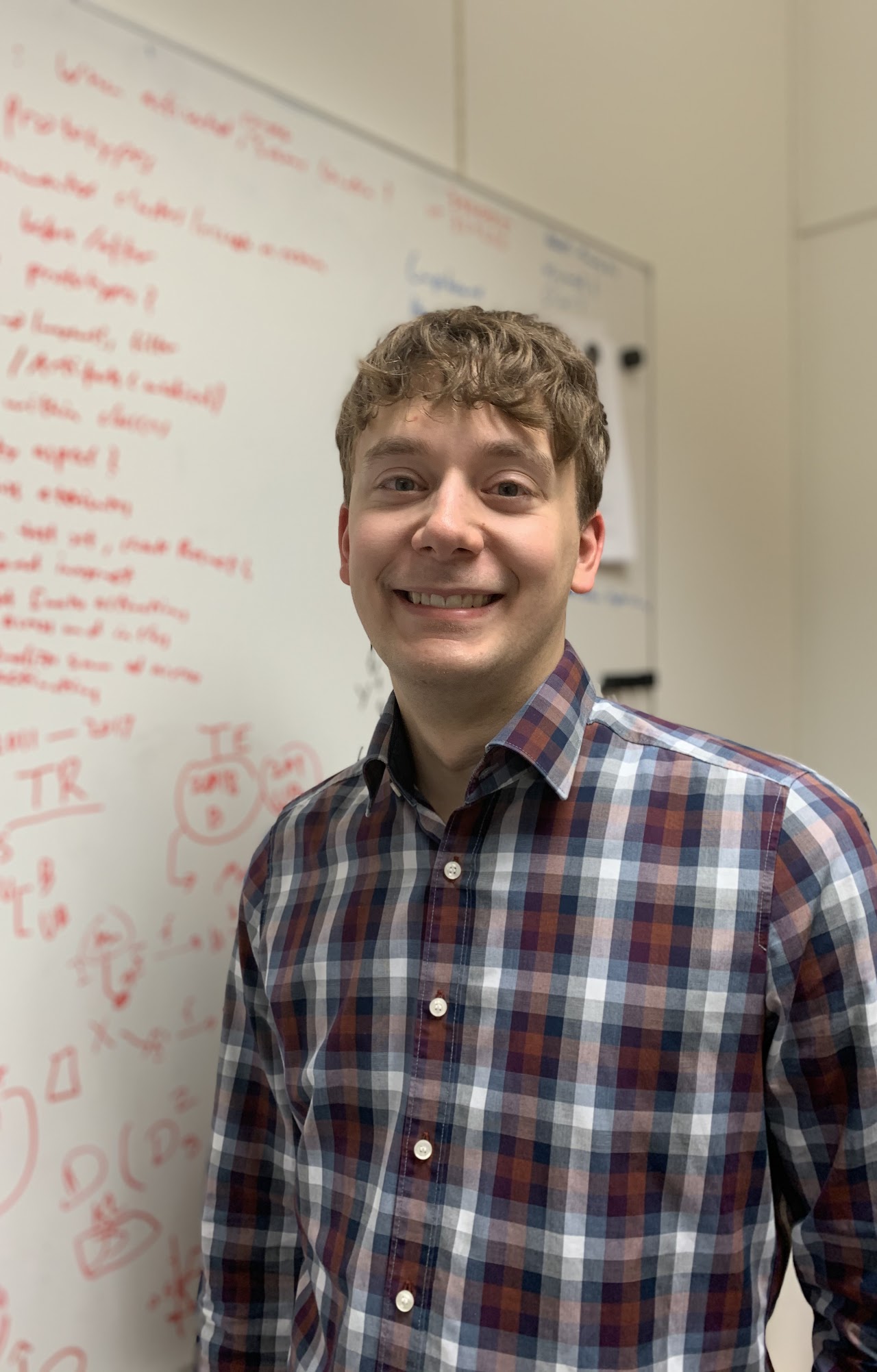}}]{Michael Kampffmeyer} is a Professor in Machine Learning at UiT The Arctic University of Norway. He is also a Senior Research Scientist II at the Norwegian Computing Center in Oslo. His research interests include medical image analysis, explainable AI, and learning from limited labels (e.g. clustering, few/zero-shot learning, domain adaptation and self-supervised learning). Kampffmeyer received his PhD degree from UiT in 2018. He has had long-term research stays in the Machine Learning Department at Carnegie Mellon University and the Berlin Center for Machine Learning at the Technical University of Berlin. He is a general chair of the annual Northern Lights Deep Learning Conference, NLDL.

\end{IEEEbiography}

\begin{IEEEbiography}[{\includegraphics[width=1in,height=1.25in,clip,keepaspectratio]{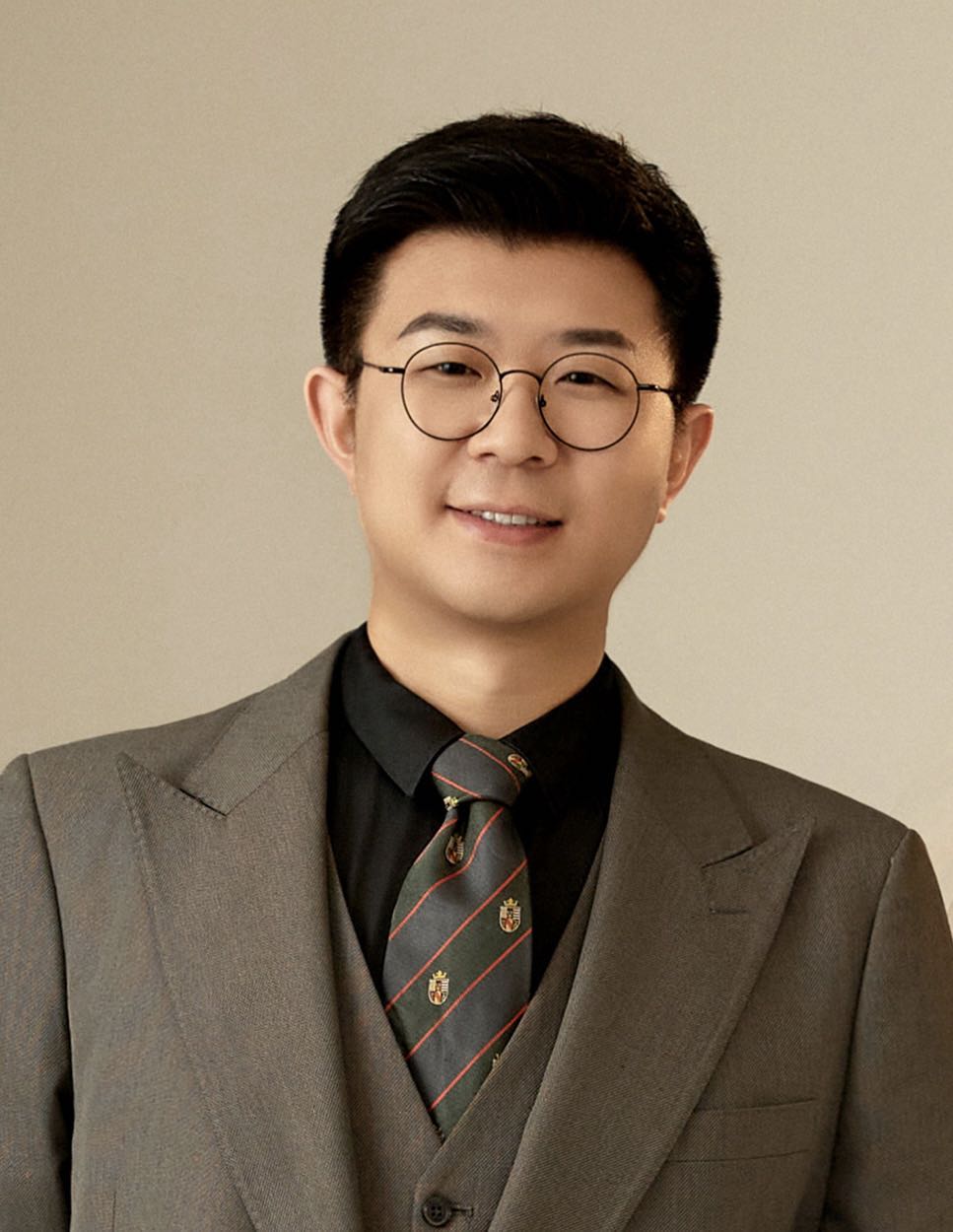}}]{Jiang Li} received his BE and ME degree from Southwest University of Science and Technology, China. Now, he is working at Meitu MT Lab. His research interests include human- centric 2D/3D synthesis.

\end{IEEEbiography}

\begin{IEEEbiography}[{\includegraphics[width=1in,height=1.25in,clip,keepaspectratio]{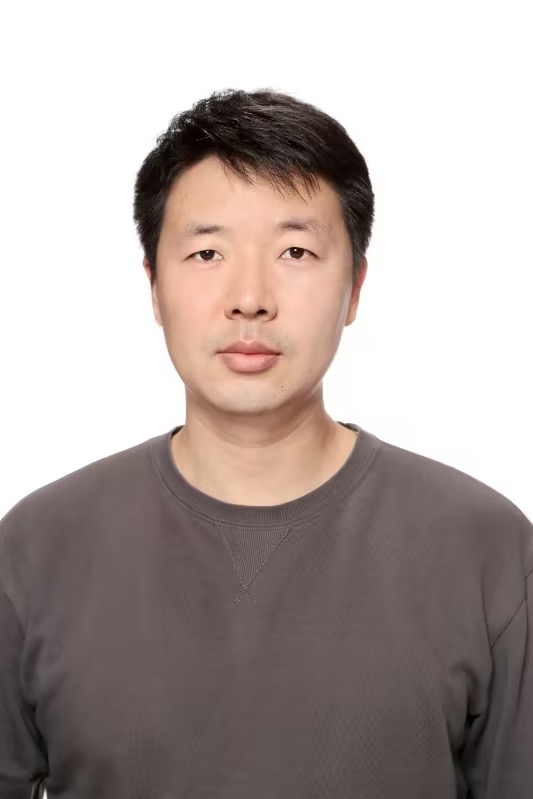}}]{Ting Liu} is currently a Senior Computer Vision Expert at Meitu Image\&Vision Lab  (Meitu, Inc.), where he leads cutting-edge research and development in segmentation, generative AI, and 3D vision. He received his M.S. in Software Engineering from Beijing Institute of Technology in 2016. His expertise spans semantic/instance segmentation, AI-driven image/video generation (e.g., inpainting, outpainting), and 3D reconstruction technologies.

\end{IEEEbiography}

\begin{IEEEbiography}[{\includegraphics[width=1in,height=1.25in,clip,keepaspectratio]{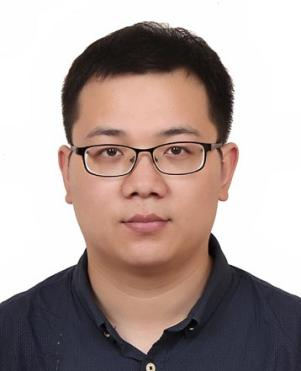}}]{Xiaochao Qu} is currently a computer vision researcher at Meitu Image\&Vision Lab  (Meitu, Inc.). He received his Ph.D. in Information Security and Management from Korea University in 2015. His research primarily focuses on computer vision, with expertise in object detection, human pose estimation, and image generation.

\end{IEEEbiography}

\begin{IEEEbiography}[{\includegraphics[width=1in,height=1.25in,clip,keepaspectratio]{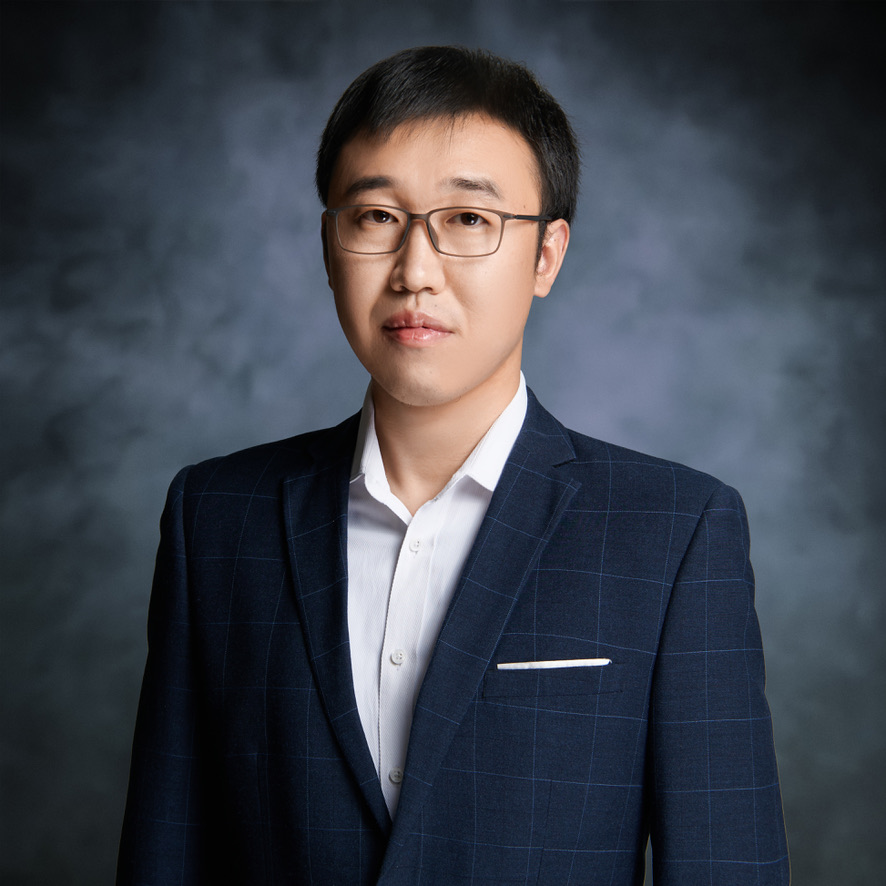}}]{Luoqi Liu} Luoqi Liu is currently Vice President of Meitu Inc and Director of Meitu Image\&Vision Lab. He is mainly working on computer vision and multimedia including face/image/video understanding, generation and  editing. He got his Ph.D. from National University of Singapore (NUS), advised by Professor Shuicheng Yan. Till now, he has published over 30 papers in top international journals and conferences, such as CVPR, NeurIPS, ECCV, ICCV, and TPAMI. He also received ACM Multimedia Best Paper Award 2013 and PREMIA Best Student Paper Gold Prize 2014.

\end{IEEEbiography}

\begin{IEEEbiography}[{\includegraphics[width=1in,height=1.25in,clip,keepaspectratio]{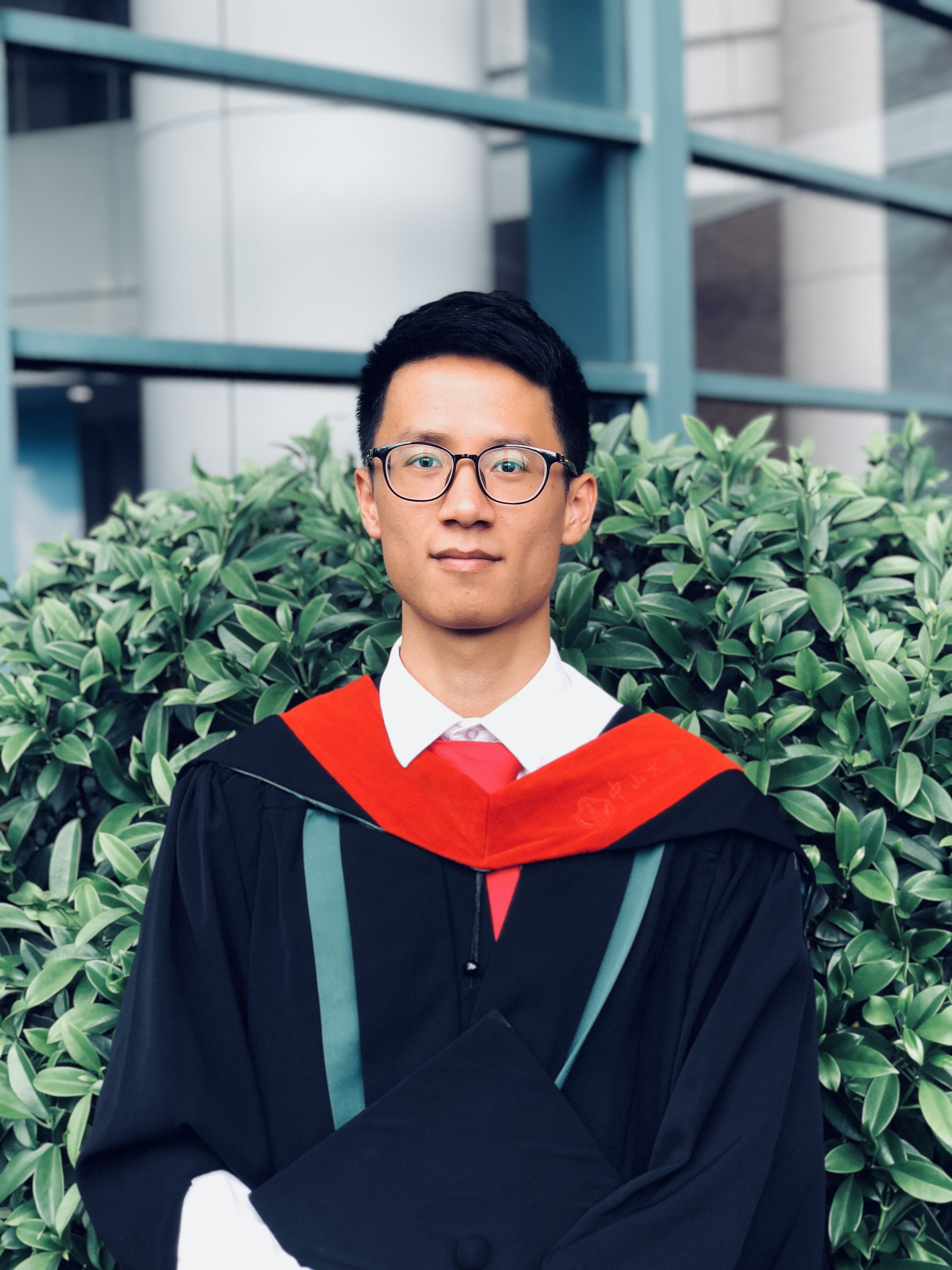}}]{Zhenyu Xie} received his BE and ME degree from Sun Yat-sen University, China. Now, he is working towards his PH.D. degree in Sun Yat-sen University. His research interests include human-centric 2D/3D synthesis.

\end{IEEEbiography}
\begin{IEEEbiography}[{\includegraphics[width=1in,height=1.25in,clip,keepaspectratio]{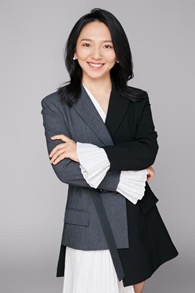}}]{Xiaodan Liang} is currently an Associate Professor at Sun Yat-sen University. She was a postdoc researcher in the machine learning department at Carnegie Mellon University, working with Prof. Eric Xing, from 2016 to 2018. She received her Ph.D. degree from Sun Yat-sen University in 2016. She has published several cutting-edge projects on human-related analysis, including human parsing, pedestrian detection, and instance segmentation, 2D/3D human pose estimation, and activity recognition.

\end{IEEEbiography}


\end{document}